\pdfoutput=1
\documentclass[11pt]{article}

\usepackage[preprint]{coling}

\usepackage{times}
\usepackage{latexsym}

\usepackage[T1]{fontenc}
\usepackage[utf8]{inputenc}

\usepackage{microtype}

\usepackage{inconsolata}

\usepackage{graphicx}
\usepackage{multirow}
\usepackage{booktabs}
\title{Predicting User Stances from Target-Agnostic Information using Large Language Models}

\author{
    Siyuan Brandon Loh, L.Z. Wong, \\
    \textbf{Prasanta Bhattacharya, Joseph Simons} \\
    Institute of High Performance Computing, \\
    Agency for Science, Technology and Research, \\
    1 Fusionopolis Way, \#16-16 Connexis, \\
    Republic of Singapore \\
    \And \And
    Hong Zhang, Wei Gao \\
    \\
    School of Computing and Information Systems, \\
    Singapore Management University, \\
    81 Victoria Street, \\
    Republic of Singapore
}

\begin{document}
\maketitle
\begin{abstract}
    We investigate the abilities of large language models (LLMs) to predict a user's stance on a target given a collection of \textit{target-agnostic} social media posts of the user. %(i.e., \emph{user-level stance prediction}).
    While we show early evidence that LLMs are capable of this task, we highlight considerable variability in the performance of the model across (i) the type of stance target, (ii) the prediction strategy and (iii) the number of target-agnostic posts supplied. Post-hoc analyses further hint at the usefulness of target-agnostic posts in providing relevant information to LLMs through the presence of both surface-level (e.g., target-relevant keywords) and user-level features (e.g., encoding users' moral values). Overall, our findings suggest that LLMs might offer a viable method for determining public stances towards new topics based on historical and target-agnostic data. 
    At the same time, we also call for further research to better understand LLMs' strong performance on the stance prediction task and how their effectiveness varies across task contexts.
\end{abstract}

\section{Introduction}
Stance is broadly defined as a social action that involves a visible expression of one’s attitudes, preferences, and viewpoints about a specific piece of information or target of interest \cite{biber2004historical,du2007stance}. With the evolution of social media, online platforms have emerged as an ideal venue for users to convey their stance on important events. Consequently, recent studies on stance detection (e.g., \citealp{mohammad2016semeval}) have attempted to analyze large corpora of social media datasets to determine whether a post, or its author, is in favor of, against, or neutral towards a target topic. 

Most prior studies on stance detection have focused on the task of detecting stance at a \emph{post}- or \emph{message}-level, with only a few studies investigating \emph{user}-level stance detection models \cite{samih2021few,Darwish_Stefanov_Aupetit_Nakov_2020,zhu2019neural}. 
Furthermore, while the extant literature has largely focused on detecting stances from posts that clearly mention or reference the stance target (i.e., \emph{target-specific} posts), the task of detecting user-level stances from posts that do not mention the stance target, and hence may or may not be related to the target (i.e., \emph{target-agnostic} posts) remains understudied \cite{aldayel2021stance,kuccuk2020stance}. In this study, we define \textit{user-level stance prediction} as this task of detecting stance from target-agnostic information posted by the user.
%\textcolor{red}{(Define stance prediction as a special case of stance detection)} 
 Stance prediction, which can be applied to new or unseen topics, is potentially of great interest and importance to organizations and policymakers. This is particularly true for contexts where limited information is available. For instance, when a new policy is announced, and the news is still unfolding, there are very few posts available directly mentioning the policy, making target-specific stance detection challenging. However, by analyzing individuals' stances on related topics or their previous posts on similar issues, we can infer their likely stance on the emerging policy. This ability to predict stances on new topics can enable organizations and policymakers to make early assessments of public opinion on emerging issues %proactively respond to emerging issues, stay ahead of public opinion, 
 and make informed decisions in response to them even when relevant data is scarce. %highlights the significance of stance prediction in real-world applications.

%Part of the reason for the lack of research in this area is that 
Despite its utility, user-level stance prediction poses significant challenges, particularly for traditional machine learning models. 
This task demands not only robust natural language understanding but also substantial background or contextual knowledge about users as well as their target topics. 
Earlier research on cross-target stance detection \cite{xu2018cross, wei2019modeling, zhang2020enhancing, liang2021target} has addressed similar issues, but these approaches are limited by their reliance on extensive target- or context-specific information. 
This highlights the need for more versatile and adaptable methods that can effectively handle new targets and contexts, making user-level stance prediction a pressing area of research.

Unlike the aforementioned traditional machine learning models, large language models (LLMs) are versatile artificial intelligence models whose pretraining has enabled them to perform various text-related tasks, including question answering, information retrieval, text summarization, and dialogue, without further fine-tuning \cite{LLM-applications}. 
In this paper, we explore the potential of LLMs to address the limitations of traditional machine learning models for user-level stance prediction. 

Our study investigates the effectiveness of LLMs in predicting a user's stance on new or unseen targets based on their authored posts which do not directly reference the target \cite{Predicting}. 
Notably, our results show that LLMs can achieve satisfactory performance with a limited volume of user-supplied, target-agnostic posts. This highlights the predictive value of target-agnostic information on social media, which has been overlooked in previous studies and datasets on stance detection.
In summary, our key contributions are four-fold:
\begin{itemize}
    \item We demonstrate early evidence that LLMs can predict user-level stances, even when the user-generated posts are target-agnostic, i.e., do not include references to the stance target. 
    \item We highlight that the performance on zero-shot stance \emph{prediction} tasks is lower than comparable stance \emph{detection} tasks where target-specific posts are available to the model. However, and notably, the performance of stance prediction quickly converges to the performance of stance detection as we increase the number of target-agnostic posts provided to the model for stance prediction.
    \item  We identify heterogeneities in LLM-based stance prediction performance based on the stance target, the prediction strategy, and the number of posts supplied.
    \item We suggest that LLMs can offer a viable method for determining public stances toward new topics based on target-agnostic data, while calling for further research to understand their effectiveness.
\end{itemize}

\section{Related Work}
The problem we address in this paper is related to prior work on cross-target stance detection, where data on a given stance target is used to train stance detection models \cite{augenstein2016stance,xu2018cross,wei2019modeling, zhang2020enhancing}, which are then adapted to infer the stance towards other targets. 
%Such studies have demonstrated the effectiveness of bidirectional conditional LSTM encoding models \cite{augenstein2016stance,xu2018cross}, self-attention mechanisms \cite{xu2018cross}, variational transfer models \cite{wei2019modeling} and graph-enhanced models \cite{zhang2020enhancing,liang2021target} in such tasks. 
%More recent work has explored the two-stage problem of target-stance extraction, where the target is discovered prior to the process \cite{li2023new}. 
Such models are designed for tasks involving \emph{related} targets or domains, often require considerable amount of \emph{target-specific} contextual information or external knowledge \cite{wei2019modeling,zhang2020enhancing}, and have exhibited efficacy on \textit{tweet-level} stance detection tasks, such as SemEval-2016 Task 6 \cite{mohammad2016semeval}. 
In contrast, we focus on a related but distinct task of predicting \textit{user-level} stances using \textit{target-agnostic} tweets which do \emph{not} include explicit references to the target.

This study draws on past work on user representation
modelling (e.g., \citealp{UserEmbedding}), which involves 
generating vector representations of users attributes  
(e.g., user-generated text, images, social network ties), 
and using them in downstream machine learning models for tasks like 
user or stance classification \cite{pennacchiotti2011machine, PredictingUser}.
While these methods have proven useful in user stance detection  \cite{PredictingUser,zhang2024enhancing}, 
they typically require the availability of ground truth labels for model training on downstream tasks. To circumvent the manual labelling of large volumes of text, recent studies have used certain tweet-level information (e.g., target-specific hashtags) to automatically label stances of tweets and users, for use in supervised and user-level stance detection models (e.g., \citealp{zhang2024doubleh,ConnectedBehaviour}).   
%As \citet{samih2021few} noted, this is a severe limitation of current stance detection models; its ability to predict a user's stance on a given target is highly predicated on the availability of a sufficiently large number of post by the user on the target of interest. 
% In the absence of \emph{target-specific} posts, manual labeling becomes imperative, which is extremely time-consuming. 
In other work, \citet{samih2021few}  performed \textit{unsupervised} stance detection by clustering users based on their tweets, including those that are not topically relevant. 
Our work shares a similar goal of attempting to carry out stance detection in the absence of \emph{target-specific tweets}. However, in contrast to their unsupervised methods, we focus on whether Large Language Models (LLMs) can make use of their contextual knowledge to draw reasonable inferences about a user's stance on a target even when target-specific information is scarce, or unavailable.

%\textcolor{blue}{This is probably overly long. We don't need many text to describe related work which ought to be concise but precise.}

Finally, our study also contributes to the broader exploration of leveraging LLMs for stance-related tasks. 
Notably, different from supervised learning models, which rely on task-specific and labeled text data, LLMs have demonstrated strong performance on certain stance detection tasks 
without requiring further optimization or fine-tuning \cite{LLM-stance,aiyappa-etal-2023-trust}.
Emerging research on zero-shot stance detection reveals that LLMs can 
effectively determine the stance of a message expressing a viewpoint towards a specific target \cite{zhang2022would,zhang2024enhancing,zhang2024llm}.
This highlights the potential of LLMs for stance-related applications, warranting further investigation into their capabilities and limitations.
In this study, we examine its capabilities on the user-level stance prediction task.

%\textcolor{blue}{Shall we need a short paragraph surveying user-level stance detection/prediction task?\\}

\begin{table*}
\centering
\begin{tabular}{p{0.15\linewidth} p{0.80\linewidth}}
\hline
\textbf{Type}                         & \textbf{Example}                                    \\ 
\hline
Target-specific              & ``\#SheGotItRight Trump and Barr don't give a shit about fascist white men threatening a woman governor when she's trying to control a pandemic.'' \\ 
\hline
Target-agnostic (related)    &``The republicans have all in all opened the biggest can of worms in history. Look at it!''                                                      \\
\hline
Target-agnostic (unrelated) & ``Ha! I do that except I use a hair band to keep it knotted on top of my head after I wash it \& put Moroccan oil in it.''                         \\ 
\hline
\end{tabular}
\caption{
Examples of tweets categorized by their relationship to the target, Donald Trump. The first example is typical in stance detection, where the target is explicitly mentioned. The second and third examples illustrate stance prediction: both examples do not explicitly mention the target, but might contain relevant cues about the target  (related), or not (unrelated).
}
\label{tab:example_tweets}
\end{table*}

\section{User-Level Stance Dataset}
%{\color{blue}{the Connected Behaviour dataset is the only dataset that provides researchers with a user's stance on selected topics, as well as both target-specific and target-agnostic tweets.}}\\
Our dataset is a sample from the Connected Behaviour (CB) dataset \cite{ConnectedBehaviour} containing tweets from 1,000 Twitter users collected between 3 April to 2 December 2020. To our knowledge, CB is the only Twitter-based and user-level stance dataset that contains tweets on multiple targets, and importantly, includes a combination of target-specific and target-agnostic tweets.
Table \ref{tab:example_tweets} presents some examples of target-specific and target-agnostic tweets from our sample. In contrast with other stance datasets such as SemEval-2016 Task 6 \cite{mohammad2016semeval}, this dataset's inclusion of user-level stances (as opposed to tweet-level stances) and target-agnostic tweets from the same users makes it uniquely suited for evaluating models on user-level stance prediction tasks. 

\textbf{Target-specific tweets} are tweets that make an explicit reference to a pre-specified topic. Each user had 2--750 target-specific tweets for each of the following three trending topics in the USA at the time:
\begin{itemize}
    \item \textbf{Donald Trump} in the 2020 election
    \item \textbf{Wearing masks} in response to COVID-19 
    \item \textbf{Racial equality} in light of racial discourse surrounding the death of George Floyd in 2020
\end{itemize}

As described in \citet{ConnectedBehaviour}, the target-specific tweets contained certain pre-determined keywords or hashtags, and  a given user's stance on each on the three topics (i.e., \emph{Support} or \emph{Against}) was determined from these target-specific tweets using a distantly supervised stance labelling procedure based on hashtags. 

\textbf{Target-agnostic tweets} are tweets that do not explicitly mention or reference the aforementioned target topics (i.e., did not contain any target-relevant keywords or hashtags). Each user had an additional 50 to 150 tweets that were target-agnostic. 
%\textcolor{red}{Foreshadow the point that target-agnostic tweets might have predictive value for two reasons: (1) they might contain target-relevant words that help the task, and (2) they might encode specific user-level attributes (e.g., moral foundations)  that are relevant to the task}
Even though target-agnostic tweets do not contain direct mentions, keywords or hashtags referencing the target, they might be conceptually relevant to the target. For example, a tweet about upholding conservative or liberal values is relevant to specific political candidates, even if the tweet does not make an explicit personal reference. 
Furthermore, target-agnostic tweets might also encode deeper user attributes, such as their beliefs and moral values, which might be useful information for the LLM, as shown in recent studies \cite{nguyen2024measuring, zhang2024enhancing}. 

On the other hand, target-agnostic tweets could also be completely unrelated to the stance target, such as the final example in Table \ref{tab:example_tweets}.
Note that we have made the distinction between related and unrelated target-agnostic tweets purely for illustrative purposes; this distinction is absent in the dataset. Indeed, part of the challenge of stance prediction is in being able to extract the `signal' in related tweets from the `noise' in the unrelated tweets.

\section{User-Level Stance Prediction Models}
In this section, we present two LLM-based stance prediction models, along with other models based on traditional ML methods. 

\subsection{LLM-based Models}
We tested two strategies for incorporating LLMs in the user-level stance prediction (ULSP) task. 
We used OpenAI's GPT-4o model\footnote{\url{https://platform.openai.com/docs/models/gpt-4o}, version 2024-05-13} which has been proven capable of carrying out a wide variety of language and reasoning tasks \cite{shahriar2024gpt4o}. 
Both strategies involved prompting the LLM using the following template:
\begin{quote}
{\small \begin{verbatim}
Read the following Tweets from a single user:
    
TWEETS
{tweets}
END OF TWEETS

Based on these Tweets, do you think this user 
supports or is against {target}?

Note that the Tweets may not mention {target}
at all. But do still try to make an educated 
guess about whether the user supports or is
against {target}.

Respond with one of these options: 
Support, Against
\end{verbatim}}
\end{quote}
\noindent with \verb"{target}" replaced by one of three targets,
``Donald Trump'', ``Wearing Masks'' and ``Racial Equality (such as the Black Lives Matter movement)'' 
and \verb"{tweets}" replaced by the user's tweets.
Responses by the LLM were mapped onto the stance labels (i.e., \emph{Support} or \emph{Against}).
Those that were not ``Support'' or ``Against'' were considered as misclassified\footnote{The user stance dataset did not have \emph{Neutral} stance labels.}. 

\paragraph{User-Level Stance Prediction - LLM}  (\emph{ULSP - LLM}). 
The first strategy involved passing a list of the user's target-agnostic tweets into the prompt. 
This enabled us to directly obtain a single user-level stance prediction from multiple tweets authored by the user.

\paragraph{User-Level Stance Prediction - LLM (pooled)} (\emph{ULSP - LLM (pooled)}).
The second strategy involved passing only one tweet
into the prompt at a time to obtain a tweet-level stance prediction.
The multiple predictions were subsequently aggregated at a user level to obtain
an average stance score for each user. 
We counted the number of tweets that resulted in \emph{Support} or \emph{Against} predictions, and returned the stance class that had a greater number of tweets.
Tweets that did not lead to \emph{Support} or \emph{Against} predictions were ignored. 

Note that both of these strategies are zero-shot methods, and do not require any ground truth labels or training.

\subsection{Traditional Machine Learning Models} 
While the primary focus of this paper is on leveraging LLMs for ULSP, we also assessed the effectiveness of non-LLM approaches to provide a comprehensive evaluation of the task. To do this, we first trained tweet-level stance prediction models. We created a tweet-level stance dataset for training by taking each user's stance label as the tweet-level label for all tweets from that user. We then generated term frequency–inverse document frequency (TF-IDF) features \cite{ramos2003tfidf}, and sentence embeddings using a pre-trained S-BERT model\footnote{\url{https://huggingface.co/sentence-transformers/all-MiniLM-L6-v2}} \cite{reimers2019sbert}, to convert each tweet into a numerical feature vector. Next, we trained a set of logistic regression and random forest classifiers (collectively called \emph{ULSP - nonLLM (pooled)}) to predict the tweet-level stance from the feature vector. %The tweet-level predictions were finally aggregated to produce a user-level stance prediction. %In the rest of this section, we detail this procedure.
%The NLP feature creation methods we used were term frequency–inverse document frequency (TF-IDF) vectorzation \cite{ramos2003tfidf}, and sentence embeddings using a pre-trained S-BERT model\footnote{\url{https://huggingface.co/sentence-transformers/all-MiniLM-L6-v2}} \cite{reimers2019sbert}.
%The machine learning models we used were logistic regression and random forest classifiers (hereafter collectively called \emph{ULSP - nonLLM (pooled)}).
The optimal regularization parameter for logistic regression and maximum depth for random forest were determined using 5-fold stratified group cross-validation, with users as groups.
All models were trained on a random sample of 50 target-agnostic tweets per user.

As with \emph{ULSP - LLM (pooled)}, we aggregated the tweet-level predictions from these models to obtain user-level predictions. Specifically, we computed a stance score for each user,
\begin{equation}
    \textsf{Stance}_i = \frac{1}{N}
    \sum\limits_{j=1}^{N} P(\emph{Support}|x_{ij})
\end{equation}
where $x_{ij}$ is the $j$-th tweet from user $i$ and 
$P(\emph{Support}|x_{ij})$ is the trained model's probability score for the \emph{Support} class given tweet $x_{ij}$ as input, and $N$ is the total number of tweets supplied. To make a final stance prediction for the user, we returned a \emph{Support} prediction if $\textsf{Stance}_i \geq T_{target}$ and an \emph{Against} prediction otherwise, where $T_{target}$ was a threshold selected to maximize the balanced accuracy of predictions on the training set.

\subsection{User-Level Stance Detection Benchmark}
In addition to applying our LLM pipelines on target-agnostic tweets (i.e., stance prediction),
we also repeated the \emph{ULSP - LLM} procedure with target-specific tweets (i.e., a traditional stance detection task),
establishing a performance ceiling that will serve as a benchmark for our models.
% The stance proportions among users in our dataset are presented in Table~\ref{tab:class_sizes}. Although we had three targets, we opted for a random split instead of stratified sampling, which would have required selecting a single target for stratification. Nonetheless, our random split yielded similar class proportions in both the training and test sets, ensuring a representative distribution of stances in each split.
\begin{table}
% \small
\centering
  \begin{tabular}{l|cc|cc}
    \hline
    \textbf{Target}&
    \multicolumn{2}{c|}{\textbf{Training set}}&
    \multicolumn{2}{c}{\textbf{Testing set}}
    \\
    &
    \emph{S}&\emph{A}&
    \emph{S}&\emph{A}
    \\
    \hline
    Donald Trump&
    138&362&
    121&379
    \\
    Wearing Masks&
    231&269&
    248&252
    \\
    Racial Equality&
    318&182&
    337&163
    \\
  \hline
\end{tabular}
  \caption{Number of users in the training and testing sets (out of 500) who support (\emph{S}) or are against (\emph{A}) each of the three stance targets based on their target-specific tweets. We used 50 tweets per user, for a total of 25,000 tweets each in the training and testing sets.}
  \label{tab:class_sizes}
\end{table}

% \textcolor{blue}{Could we also add the \# of posts of each of the dataset? That can give readers a specific sense on how large the dataset is.} \textcolor{green}{There are 500 users for each dataset, with the same number of tweets (50 target-agnostic tweets per user). Don't know if we need to mention that. But I added "(out of 500)" above.}
%\textcolor{red}{In section 3, you mentioned that there are 2-750 target-specific tweets and 50-150 target-agnostic tweets for each user. Then, it's not the same number of tweets per user, right?} 
%\textcolor{green}{We don't use the target-specific tweets though. And we also don't use the full number of target-agnostic tweets. Only 50 per user. So if we put number of tweets, it's just all the above numbers, times 50. I'm not opposed to putting the number of tweets, but I don't know where it would fit into this table. And we've already mentioned that in the text. The original purpose of the table was to show the class size proportions, which is reflected in the number of users rather than number of tweets.}
%\textcolor{red}{That's fine as it's mentioned in the text. I just assumed different number of posts are associated with each user. Anyway it would be better to tell the total number of posts in the training set and test set (in the text as well.} 
%\textcolor{green}{Ok I have mentioned 25,000 tweets for the training and test set in the text.}

\begin{table*}[h!]
% \small
\centering
\begin{tabular}{p{0.15\linewidth}|p{0.4\linewidth}|p{0.1\linewidth}|p{0.1\linewidth}|p{0.1\linewidth}}
\hline
\textbf{Target}&
\textbf{Model}&
\textbf{F1 \newline(Support)}&\textbf{F1 (Against)}&\textbf{Balanced Accuracy}
\\
\hline
Donald Trump&
ULSP - LLM&
\textbf{0.784}&\textbf{0.936}&\textbf{0.850}
\\
&
ULSP - LLM (pooled)&
0.587&0.907&0.711
\\
&
ULSP - NonLLM (TFIDF+LogReg)&
0.625&0.869&0.760
\\
&
ULSP - NonLLM (SBERT+LogReg)&
0.520&0.767&0.692
\\
&
ULSP - NonLLM (SBERT+RanFor)&
0.444&0.584&0.613
\\
\hline
Wearing Masks&
ULSP - LLM&
\textbf{0.746}&\textbf{0.559}&\textbf{0.680}
\\
&
ULSP - LLM (pooled)&
0.733&0.575&0.674
\\
&
ULSP - NonLLM (TFIDF+LogReg)&
0.639&0.546&0.599
\\
&
ULSP - NonLLM (SBERT+LogReg)&
0.627&0.540&0.589
\\
&
ULSP - NonLLM (SBERT+RanFor)&
0.634&0.473&0.569
\\
\hline
Racial Equality&
ULSP - LLM&
\textbf{0.880}&\textbf{0.667}&\textbf{0.749}
\\
&
ULSP - LLM (pooled)&
0.862&0.638&0.733
\\
&
ULSP - NonLLM (TFIDF+LogReg)&
0.752&0.606&0.702
\\
&
ULSP - NonLLM (SBERT+LogReg)&
0.768&0.613&0.709
\\
&
ULSP - NonLLM (SBERT+RanFor)&
0.698&0.571&0.666
\\
\hline
\end{tabular}
\caption{Performance metrics (F1 and balanced accuracy) for stance prediction with 50 target-agnostic tweets. Best scores for each target and metric are in bold. }
\label{tab:metrics}
\end{table*}

\section{Experiments and Results}
In this section, we begin by presenting the performance of different stance prediction models, and illustrate the effect of varying the numbers of tweets provided to the model. We then present some secondary analyses that were carried out in an attempt to uncover what the LLMs might be picking up on when making predictions, and to mitigate bias in making these predictions.

\subsection{Cross-validation for Non-LLM Methods}
Unlike the case for zero-shot LLM stance prediction or detection, the non-LLM models (i.e., logistic regression and random forest) described above needed to be trained on labelled data. 
To that end, we randomly divided the 1,000 users into training and testing sets, each comprising 500 users. The stance proportions for the train and test users are presented in Table~\ref{tab:class_sizes}.  Only the non-LLM models were trained on data from the 500 training set users. The 500 testing set users were used to evaluate the zero-shot LLM prediction and detection models as well as the trained non-LLM models.
We used 50 target-agnostic tweets per user for training and testing, for a total of 25,000 tweets in the training set and testing set.

\subsection{Main Results of Performance}
Table \ref{tab:metrics} presents the performance results of the various methods when making predictions based on a target-agnostic tweets per user. As shown in the table, \emph{ULSP - LLM}  outperformed all models across all stance targets and performance metrics. Apart from the target ``Donald Trump'', 
\emph{ULSP - LLM (pooled)} prediction also performed better than all 
\emph{ULSP - nonLLM} models, but not as well as \emph{ULSP - LLM} .

We also observe considerable variability in model performance across stance targets. Across methods, predictions for ``Donald Trump'' achieved the highest accuracy, followed by ``Racial Equality'' and ``Wearing Masks'', indicating target-specific differences in performance.

\begin{figure*}[h!]
  \includegraphics[width=\linewidth]{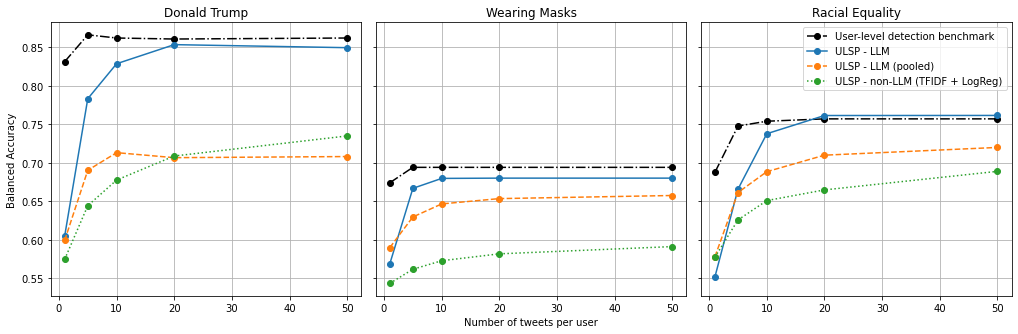}
  \caption{Performance of stance prediction models across three stance targets. Balanced accuracy of \emph{ULSP - NonLLM (TFIDF + LogReg)},
  \emph{ULSP - LLM (pooled)}, and \emph{ULSP - LLM} across ``Donald Trump'', ``Wearing Masks'', and ``Racial Equality'' as a function of tweets per user. Stance detection performance (\emph{User-level detection benchmark}) is included for comparison.}
  \label{fig:balanced_accuracy}
\end{figure*}

Across all models, increasing the number of tweets generally improved the balanced accuracy, as shown in Figure \ref{fig:balanced_accuracy}. As expected, stance \emph{detection} generally achieved the highest balanced accuracy with as few as 5 target-specific tweets. This corroborates findings from \cite{samih2021few} that only a few tweets are sufficient for stance detection if the tweets are topically relevant (i.e. target-specific). Notably, the performance of \emph{ULSP - LLM} eventually caught up to that of stance \emph{detection} given sufficient tweets (10 or 20), demonstrating that LLMs can successfully carry out stance prediction given a relatively small number of target-agnostic tweets (10 or 20), at a level that matches stance detection when target-specific tweets are available.

While the other models also improved as the number of tweets increased, their performance paled in comparison to \emph{ULSP - LLM}. Surprisingly, while \emph{ULSP - LLM (pooled)} performed better than \emph{ULSP - nonLLM (pooled)} for the ``Wearing Masks'' and ``Racial Equality'' targets, it performed worse for the ``Donald Trump'' target. We investigate the reason for this issue in the following section. %will revisit this issue in Section \ref{sec:threshold}.

\subsection{\emph{ULSP - LLM (pooled)} with Thresholding}
\label{sec:threshold}
\begin{figure*}[htp!]
  \includegraphics[width=\linewidth]{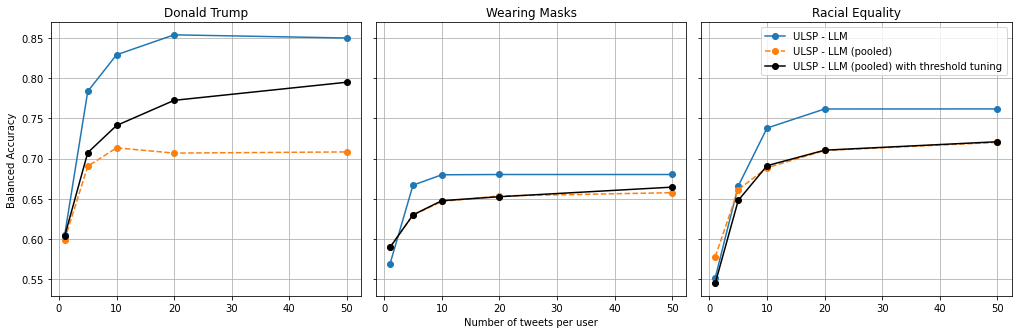}
  \caption{Effect of threshold tuning on the performance of \emph{ULSP - LLM (pooled)}, with  \emph{ULSP - LLM} included for comparison.}
  \label{fig:tuning_threshold}
\end{figure*}
In this section, we perform an investigation of the discrepancy in performance between \emph{ULSP - LLM (pooled)} and \emph{ULSP - LLM} . Unlike target-specific tweets, most target-agnostic tweets lack sufficient information about a user's broader beliefs or preferences when considered in isolation. 
We hypothesize that this makes \emph{ULSP - LLM (pooled)} more prone to error, which can then introduce biases when aggregated to make inferences about users.
% For stance prediction, the LLM's performance depends on the specific methods and techniques employed; across all targets, only the \textbf{user-level LLM's} performance converged to stance detection performance as the number of tweets increased. While the exact cause of this performance difference is unclear, we hypothesize that the contextual information provided by each tweet in the \textbf{user-level LLM}, where most target-agnostic tweets are considered in conjunction with one another, contributes to the improved performance. 

Recall that \emph{ULSP - LLM (pooled)} gives a \emph{Support} prediction when
\begin{equation}
\frac{1}{N} \left(\textsf{\# Support} - \textsf{\# Against} \right) \geq 0,
\end{equation}
where $N$ is the number of tweets supplied. However, with the availability of training data, this threshold can be further tuned to optimize the accuracy on the training set.
This can also help to counter systematic biases in the tweet-level predictions, e.g., if the LLM tends to predict more of one particular class.

In Figure \ref{fig:tuning_threshold}, we see that this threshold tuning led to significant improvements for the ``Donald Trump'' target but minimal change for the other two targets, suggesting that the original tweet-level predictions for ``Donald Trump'' might have exhibited systematic bias that were countered by the threshold tuning. 

Despite the improvement, however, \emph{ULSP - LLM (pooled)} remained less accurate than \emph{ULSP - LLM}.
We hypothesize that when passing multiple tweets together, instead of one at a time, LLMs are able to utilize the more relevant tweets as context to better understand the ambiguous ones, thereby enhancing their predictive capabilities.

\subsection{Relevance of Target-agnostic Tweets}

\begin{figure}[h!]
  \includegraphics[width=\linewidth]{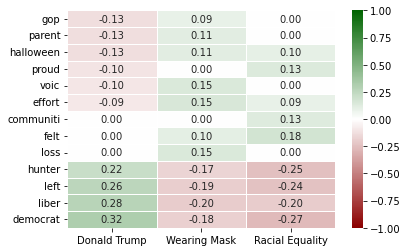}
  \caption{Biserial correlation coefficients between selected stemmed terms and stance on 3 targets.}
  \label{fig:correlations}
\end{figure}

Our results highlight that predictive value of utilizing target-agnostic tweets in predicting users' stance towards an unseen target. As discussed earlier, we believe that this might be driven by two characteristics of the target-agnostic tweet - (a) the presence of target-relevant keywords in the tweet and/or (b) the encoding of deep user characteristics. 

\paragraph{Target-relevant keywords.} Target-agnostic tweets might be predictive of targets if the tweets do not directly reference the target, but contain certain keywords or terms that might be conceptually related to the target. To test if this was true, we computed correlations between TF-IDF features and stance classes. 
Figure \ref{fig:correlations} shows selected terms that correlated strongly (positively or negatively) with stances towards each target. This illustrates that target-agnostic tweets can still be related to the target, even though they may not explicitly mention the target.

However, the performance gap between \emph{ULSP - LLM} and the \emph{ULSP - nonLLM} models, which make use of these terms, suggests that LLMs are able to leverage contextual information beyond the presence of such related terms.

\begin{figure}[h!]
  \includegraphics[width=\linewidth]{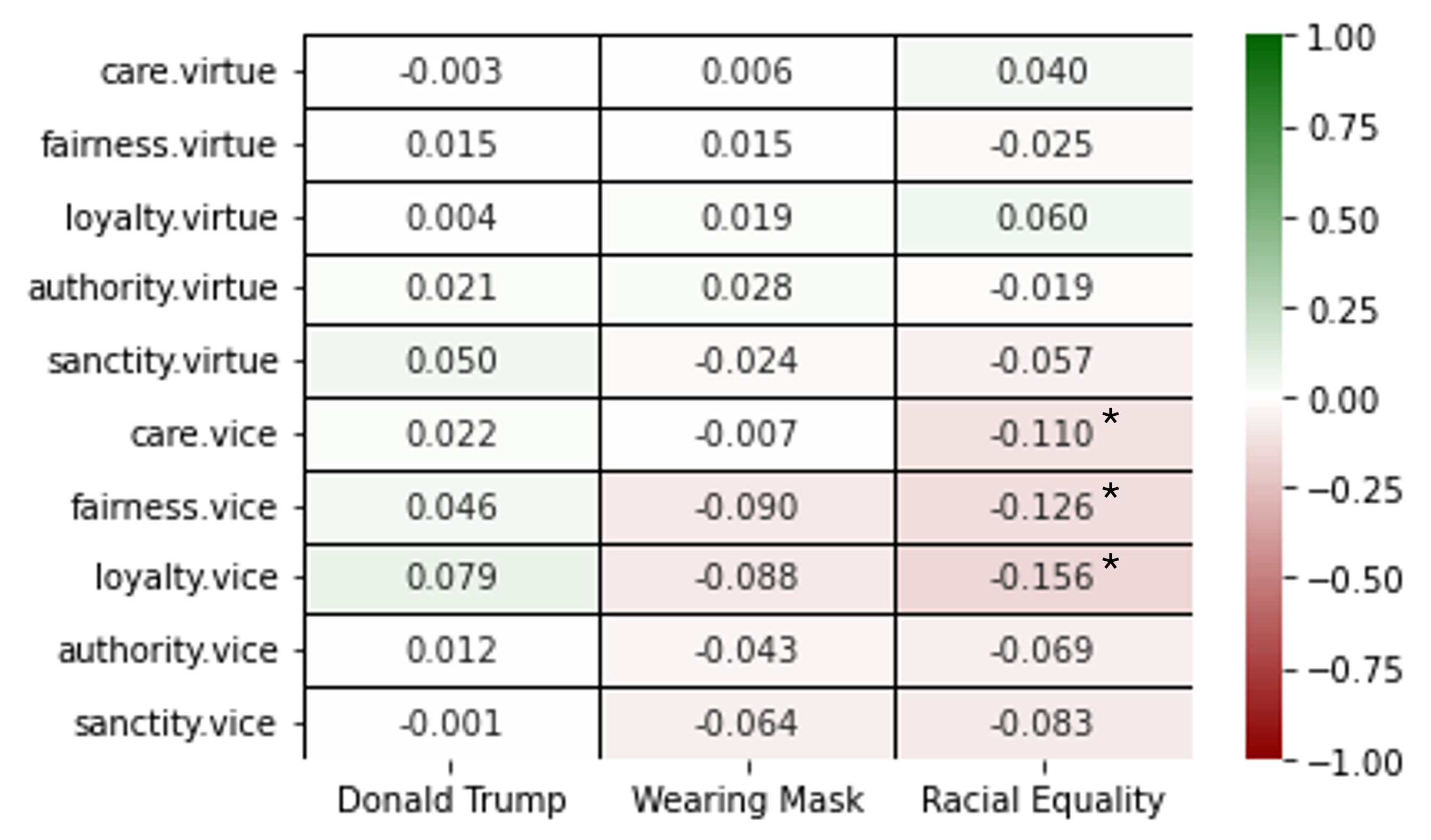}
  \caption{Biserial correlation coefficients between eMFD scores of \textit{target-agnostic tweets} and stance on three targets. Cells with an asterisk (*) indicate \( p < .01 \).}
  \label{fig:eMFD}
\end{figure}

\paragraph{Deep user characteristics.} Beyond the presence of informative keywords, another factor driving the predictive value of target-agnostic tweets could be the LLM's ability to make deep inferences about an individual (e.g., their beliefs and values systems) based on a collection of their past tweets which provides useful user-level context to the model. For instance, a user who expresses a favourable attitude towards government regulations on combating climate change might value authority and care over other moral considerations. As such, this person might 
hold a positive stance towards COVID-19 related policies on vaccine and mask mandates since this aligns with their broader values. Recent studies have shown that LLMs are able to extrapolate beyond the immediate context of the message to make such inferences about users \cite{nguyen2024measuring,zhang2024enhancing}, including but not limited to their stance on unseen targets. 

In order to test this conjecture, we further conducted a post-hoc analysis to examine if users who differed in their stance on a given topic reflected systematic differences in individual-level factors (e.g., moral considerations) as inferred from their target-agnostic tweets. In particular, we wanted to examine if these tweets differed along some dimension that can loosely be mapped to a user's broader belief/value system. To this end, we concatenated all tweets from the same user 
and then extracted their moral foundation  
scores from the concatenated tweets using the extended Moral Foundation Dictionary (eMFD) provided by \citet{hopp2021extended}.

We observed a significant negative relationship in the 
"care.vice," "loyalty.vice," and "fairness.vice" scores 
of target-agnostic tweets and stance on ``Racial Equality''. 
This indicates that supporters of ``Racial Equality'' scored 
lower on these three dimensions, 
reflecting a lower tendency to generate content associated with 
cruelty, injustice, and betrayal compared to their counterparts 
who oppose ``Racial Equality''. 
The eMFD scores did not significantly differ among users with different stances toward ``Donald Trump'' or ``Wearing Masks''.

Altogether, our post-hoc analyses provided some evidence 
that users who differed on their stance towards a given target 
also expressed themselves differently on other topics, 
as measured by eMFD scores or stemmed terms. 
While this does not conclusively explain the underlying mechanisms behind LLMs performance on stance prediction, they suggest that subtle cues within tweets can be leveraged to make predictions about user attributes, even in the absence of explicit information regarding those attributes.

\section{Conclusion}
In this work, we have demonstrated that LLMs are capable of carrying out user-level stance prediction given sufficient target-agnostic tweets, to a level comparable to stance detection from target-specific tweets.
Our experiments show that LLMs outperform traditional non-LLM methods, and supplying multiple tweets to an LLM all at once (\emph{ULSP - LLM}) works better than supplying them individually (\emph{ULSP - LLM (pooled)}, even with threshold tuning).
We have also attempted to demonstrate that LLMs might be picking up on target-relevant keywords and deep user characteristics that might be present in target-agnostic tweets.

However, it remains unclear how increasing the number of tweets in the user-level strategy enhances performance. Does the improvement stem from a gradual increase in information about the user, or is it due to a higher likelihood of including particularly diagnostic tweets that inform the prediction? These questions highlight the need for further investigation into the mechanisms underlying LLMs' performance on user-level stance prediction and related tasks.

\section{Limitations and Future Work}
Our study was limited by the availability of suitable datasets. 
As far as we know, the CB dataset is the only publicly available resource that provides researchers with a user's stance on selected topics, as well as both target-specific and target-agnostic tweets. 
% Most stance-related datasets, however, only contain target-specific tweets, making them unsuitable for the task of user-level stance prediction. 
The availability of additional datasets that include both target-specific and target-agnostic tweets will  
enable more comprehensive research in user-level stance prediction.

Our LLM prompting strategy was also relatively simple. Future work can look into more sophisticated prompting methods, such as getting the LLM to carry out ``reasoning'' before making a stance prediction (e.g., following \citet{COT}), or providing additional contextual information about the stance target (e.g., definitions or descriptions of the stance target). Furthermore, this work only tested the use of one specific LLM; future work could look into comparing the performance of different LLMs.

\section*{Acknowledgments}
This research is supported by the SMU-A*STAR Joint Lab in Social and Human-Centered Computing (SMU grant no.: SAJL-2022-CSS02, SAJL-2022-CSS003).\\
This research is supported by A*STAR (C232918004, C232918005).
\bibliography{coling}

\end{document}